\documentclass[sigconf, nonacm]{acmart}

\AtBeginDocument{%
  \providecommand\BibTeX{{%
    \normalfont B\kern-0.5em{\scshape i\kern-0.25em b}\kern-0.8em\TeX}}}

\usepackage{amsmath}
\usepackage{bm}
\usepackage{multirow}
\usepackage{caption}
\usepackage{graphicx, subfig}
\usepackage{booktabs}
\usepackage[ruled, linesnumbered]{algorithm2e}

\begin{document}

\title{CAP: Co-Adversarial Perturbation on Weights and Features for Improving Generalization of Graph Neural Networks}

\author{Haotian Xue}
\affiliation{Jilin University \country{}}
\email{xueht20@mails.jlu.edu.cn}

\author{Kaixiong Zhou}
\affiliation{Rice University \country{}}
\email{Kaixiong.Zhou@rice.edu}

\author{Tianlong Chen}
\affiliation{
  University of Texas at Austin \country{}
}
\email{tianlong.chen@utexas.edu}

\author{Kai Guo}
\affiliation{
  Jilin University \country{}
}
\email{guokai20@mails.jlu.edu.cn}

\author{Xia Hu}
\affiliation{
  Rice University \country{}
}
\email{xia.hu@rice.edu}

\author{Yi Chang}
\affiliation{
  Jilin University \country{}
}
\authornote{Corresponding Authors}
\email{yichang@jlu.edu.cn}

\author{Xin Wang}
\affiliation{
  Jilin University \country{}
}
\authornotemark[1]
\email{xinwang@jlu.edu.cn}

\begin{abstract}
Despite the recent advances of graph neural networks (GNNs) in modeling graph data, the training of GNNs on large datasets is notoriously hard due to the overfitting. Adversarial training, which augments data with the worst-case adversarial examples, has been widely demonstrated to improve model's robustness against adversarial attacks and generalization ability. However, while the previous adversarial training generally focuses on protecting GNNs from spiteful attacks, it remains unclear how the adversarial training could improve the generalization abilities of GNNs in the graph analytics problem. In this paper, we investigate GNNs from the lens of weight and feature loss landscapes, i.e., the loss changes with respect to model weights and node features, respectively. We draw the conclusion that GNNs are prone to falling into sharp local minima in these two loss landscapes, where GNNs possess poor generalization performances. To tackle this problem, we construct the co-adversarial perturbation (CAP) optimization problem in terms of weights and features, and design the alternating adversarial perturbation algorithm to flatten the weight and feature loss landscapes alternately. Furthermore, we divide the training process into two stages: one conducting the standard cross-entropy minimization to ensure the quick convergence of GNN models, the other applying our alternating adversarial training to avoid falling into locally sharp minima. The extensive experiments demonstrate our CAP can generally improve the generalization performance of GNNs on a variety of benchmark graph datasets.
\end{abstract}





\maketitle

\section{Introduction}

Graph neural networks (GNNs) have emerged as promising deep learning tools to analyze networked data, such as social networks~\cite{gnn4social, deepinf}, academic networks~\cite{planetoid}, and molecular graphs~\cite{molecular1}. Based on spatial graph convolutions, GNNs learn the representation of each node by aggregating the representation embeddings of node itself and its neighbors recursively~\cite{graphsage}. Despite the superior performance in graph analytics, the standard training of GNNs is notoriously challenging, and usually suffers from the vulnerability to adversarial samples \cite{attong, attongraph} and the overfitting issue \cite{flag, zhou2021adaptive}.

Among the recent emergence of promising techniques, adversarial training has been empirically demonstrated to achieve the superiority in terms of adversarial robustness as well as generalization for many real-world applications, such as the visual recognition \cite{advex, afan} and language modeling \cite{nlmat, freelb}. However, in the graph analytics, the existing adversarial training based methods \cite{robustgnn, robgcn, robgnntrans} generally focus on enhancing the network robustness against the handcrafted adversarial examples, instead of improving the generalization ability. Specifically, the large-scale graphs often contain large volumes of out-of-distribution testing nodes \cite{ogb}, which possess the distinct features and neighbor structures to those in training sets. This makes us pose the following research question: Whether the adversarial training could improve the generalization performance and boost the test accuracy in GNNs?

\begin{figure*}
    \includegraphics[width=\textwidth]{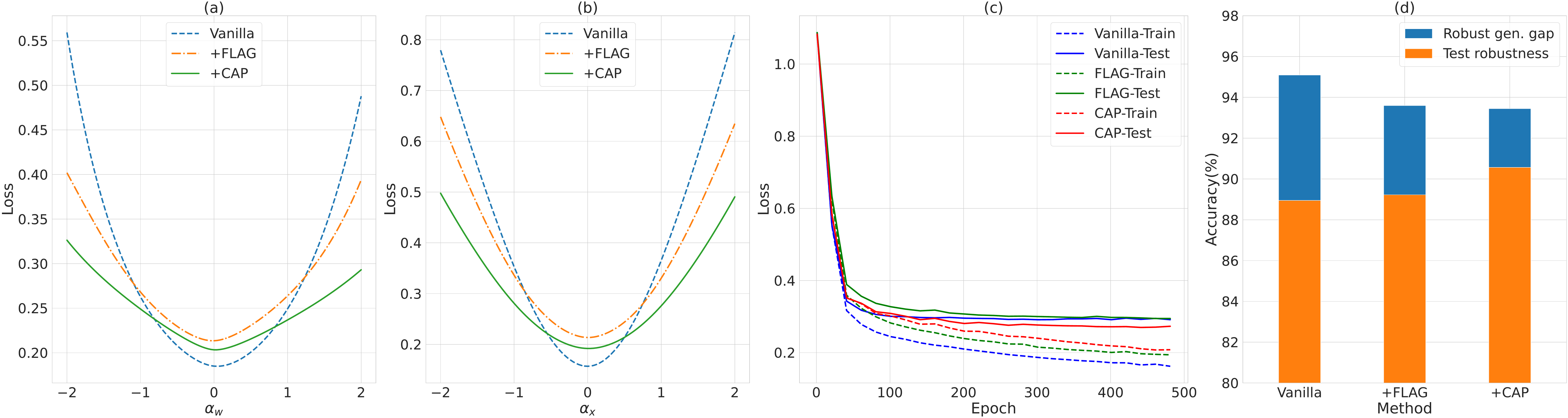}

    \caption{Over GCN backbone on \emph{Pubmed} dataset, the results of vanilla training, FLAG, and CAP in terms of (a) weight loss landscape, (b) feature loss landscape, (c) training and testing losses, and (d) training and testing accuracies.}
    \label{fig:lossland}
\end{figure*}

To answer the question, we first study the standard training of vanilla GNNs from the lens of loss landscape and classification accuracy in Figure \ref{fig:lossland}, and draw the following findings. First, training on \emph{Pubmed}~\cite{pubmed}, graph convolutional network (GCN) \cite{gcn} converges to sharp local minima in both the weight and feature loss landscapes (see Figure \ref{fig:lossland} (a) and (b)). 
Such sharp minima are notoriously known to closely correlate with the poor generalization abilities of deep neural networks~\cite{gapsm, smgendnn, sam}.
Second, the generalization gap (i.e., the gap between training and testing accuracies in \ref{fig:lossland} (d))  is the highest for the vanilla training, which directly indicates the unpleasant generalization and the overfitting on testing and training data, respectively. While some of the adversarial training methods start to flatten the feature loss landscape in vision and natural language models, a few of them instead explore to smooth the weight loss landscape~\cite{awp}. Given the complex couple sharpness in GNNs, it still remains unclear how to leverage the adversarial training to simultaneously optimize the two loss surfaces and comprehensively improve the generalization ability.

Towards bridging the gap, we propose the co-adversarial perturbation (CAP) to explicitly regularize the weight and feature loss landscapes for GNNs. Typically, we formulate the co-adversarial training objective to minimize the maximum training loss within a couple regions of model weights and node features. To efficiently solve the co-adversarial problem, we decouple the training objective and propose the alternating adversarial perturbations: one step injecting the adversarial weight perturbation and training GNN models, and another step calculating the adversarial feature perturbation for each node to update GNNs. While the weight perturbation globally works on all the node samples and optimizes the model-wise worst loss, the feature perturbation improves the local generalization by minimizing the node-wise worst loss. They are complementary to each other to iteratively flatten both the weight and feature loss landscapes. Furthermore, we observe the vanilla adversarial training often disturbs the efficient convergence at the initial training stage of GNNs, which even leads to worse generalization performances. We thus design a two-stage training scheme, where the first stage conducts the standard training to isolate the initial model convergence from the adversarial noise, and the second stage uses CAP to avoid model converging to sharp local minima. We summarize our contributions as follows:

\begin{itemize}
    \item We deliver the first step to study the weight and feature loss landscapes for GNNs, and identify the fact GNNs converge to locally sharp minima and have poor generalization performances on the node classification tasks.
    \item We propose CAP to jointly perturb the model weights and node features, and flatten the worst-case adversarial loss in the loss landscape. We then propose the alternating adversarial  perturbations of weights and features to efficiently solve the co-adversarial problem. A two-stage training scheme is applied to ensure the quick convergence of GNNs at the initial training stage, which moves GNNs toward the reliable regions in the loss landscape generally accompanied with smaller losses.
    \item We conduct extensive experiments by applying CAP for different GNN backbones evaluated on various benchmarks, and show that CAP could generally improve the testing accuracy and shrink the generalization gap.
\end{itemize}

\section{Related Work}
\paragraph{\textbf{Graph neural networks}} \citet{bruna2013spectral} firstly propose a Convolutional Neural Network in graph domain based on the spectrum of the graph Laplacian.
Further, \citet{gcn} propose a simple layer-wise propagation rule which restricts the filters to operate in one-step neighbors around each node, and illustrate how it can be motivated from a first-order approximation of spectral graph convolutions. \citet{gat} introduce graph attention networks leveraging masked self-attentional layers. They are under transductive settings with full-batch training, which means they are not suitable for large-scale graphs.
For the above limitations, ~\citet{graphsage,yelp} raise mini-batch training algorithms on graphs in performing graph sampling methods.
There are some attempts to build deeper GNNs in \cite{dirichletgnn, diffnorm}.

\paragraph{\textbf{Adversarial training}} Adversarial training is a widely used countermeasure for adversarial attacks on image data. The key of adversarial training is mixing adversarial examples into the clean training set such that the trained model can correctly classify the future adversarial examples. Adversarial samples are firstly demonstrated by \citet{advsamples}. To defense the interference of adversarial samples on the model, many defensive approaches have been developed such as input defensive distillation~\cite{distill}, feature squeezing~\cite{featsqueeze}, gradient regularization~\cite{inputgrad, datagradreglarization}, and adversarial training~\cite{fgsm, fgm, pgd, freeAT}. \citet{obfuscated} show that adversarial training is the most effective method that demonstrates moderate robustness and has thus far not been comprehensively attacked.

\paragraph{\textbf{Adversarial training on graphs}} Owing to the non-Euclidean structure of the graph data, the adversarial training on graphs is also different. Specifically, the graph data is generally composed of the adjacency matrix $\mathbf{A}$ and the node feature $\mathbf{X}$, so the perturbation object that is used to craft the adversarial sample is alternative. \citet{atonadj} suggest dropping edges randomly in adversarial training to generate perturbations on the adjacency matrix $\mathbf{A}$. Furthermore, \citet{topoatt} introduce projection gradient descent (PGD) to generate perturbations instead of dropping edges randomly. \citet{bvat} propose batch virtual adversarial training algorithm which smooths the output distribution of graph-based classifiers. \citet{flag} regard adversarial training as a method of data augmentation and propose FLAG which iteratively augments node features with gradient-based adversarial perturbations during training phase. \citet{latent} raise a latent adversarial training method that injects perturbations on the hidden layer.

\paragraph{\textbf{The connection between the loss landscape and generalization}}
The relationship between the geometry of the loss landscape — specifically, the flatness of minima — and generalization has been intensively investigated from theoretical and experimental points of view~\cite{sam, nonvacuous, smgendnn}.
~\citet{gapsm} and~\citet{smgendnn} observe that the lack of generalization ability is due to the fact that the loss function of deep neural networks is tend to converge to sharp local minima. In adversarial training, there are two types of loss landscape: input loss landscape and weight loss landscape. The first reflects the change of loss in the vicinity of training examples. In this view, Adversarial training explicitly flattens the loss landscape by training on adversarial examples. The last is the loss change with respect to the weight. \citet{sam} propose Sharpness-aware Minimization (SAM) that explicitly smooths the weight loss geometry during model training. \citet{cvcoadv} attempt to incorporate SAM into adversarial training. \citet{awp} discover a relation between the flatness of the weight loss landscape and the robust generalization gap in CV, while offer Adversarial Weight Perturbation (AWP) as a way to explicitly regularize the flatness of the weight loss landscape.

\section{Loss Landscape Analysis}
In this section, we start by introducing the notations and GNNs, and then analyze the weight and feature loss landscapes, which lead to deep understanding of the generalization ability of GNNs.

\subsection{Notations and GNNs} We denote matrices with boldface capital letters (e.g. $\bm{X}$), vectors with boldface lowercase letters~(e.g., $\bm{x}$) and scalars with lowercase alphabets~(e.g., $x$). An undirected graph is represented by $G=(\mathcal{V}, \mathcal{E})$, where $\mathcal{V} = \{v_i\}$ and $\mathcal{E} = \{(v_{i}, v_{j})\}$ denote the node and edge sets, respectively. Let $\bm{X} \in \mathbb{R}^{n \times d}$ denote the node feature matrix, where the $i$-th row is the corresponding $d$-dimensional feature vector of node $v_i$. The adjacency matrix is defined as $\boldsymbol{A} \in \mathbb{R}^{n \times n}$, which associates each edge $(v_{i}, v_{j})$ with its element $A_{i j}$; and $\bm{D}$ is the degree matrix. Let $ \tilde{\bm{A}}:= \bm{A} + \bm{I}_n$ and $\tilde{\bm{D}} := \bm{D} + \bm{I}_n$ be the adjacency and degree matrices of the graph augmented with self-loops. The normalized adjacency matrix is given by $\hat{\bm{A}} := \tilde{\bm{D}}^{-\frac{1}{2}}\tilde{\bm{A}}\tilde{\bm{D}}^{-\frac{1}{2}}$, which is widely used for spatial neighborhood aggregation in GNN models.

We use GCN~\cite{gcn} as a typical example to illustrate the node representation learning. The forward inference at the $l$-th layer of GCN is formally defined as:

\begin{equation}
\bm{H}^{(l)}=\sigma(\hat{\bm{A}} \bm{H}^{(l-1)} \bm{W}^{(l)}),
\label{eq:GCN}
\end{equation}
where $\bm{H}^{(l)}$ denotes the node embedding matrix at the $l$-th layer; $\bm{H}^{(0)}$ is given by $\bm{X}$;
$\sigma(\cdot)$ is the nonlinear activation function, such as $\mathrm{ReLU}$; $\boldsymbol{W}^{(l)} \in \mathbb{R}^{d \times d}$ is the linear transformation matrix. It is observed that graph convolutions consist of two key steps: the spatial neighbor aggregation based upon matrix $\hat{\bm{A}}$ and the feature transformation with matrix $\bm{W}^{(l)}$. Let $L$ denote the depth of the model. The output embedding of node $v_i$, i.e., $\bm{h}^{(L)}_i$ at the $i$-th row of $\bm{H}^{(L)}$, is usually applied to predict node labels.

\subsection{Why GNNs Generalize Poorly}
Considering the node classification task in graph analytics, the vanilla training based on cross-entropy minimization often leads to over-confident prediction on the training data and poor generalization to the testing data~\cite{zhou2021adaptive}. It is also reported that the vanilla training of GNNs is sensitive to overfitting~\cite{chen2021bag, flag}. These all point to the optimization problems. We first analyze the loss
landscapes, and shred deep insight by building the close connection between sharp local minima and worse generalization performance.

\paragraph{\textbf{Weight and feature loss landscapes}} It has been widely demonstrated that the model convergence to a flat region in the loss landscape, whose curvature is small, could ameliorate the generalization of deep neural networks \cite{gapsm, smgendnn, sam}. We characterize the training process of GNNs with both the weight and feature loss landscapes. They are defined by the loss changes when moving the weights and features along random directions with scalar magnitudes:
\begin{equation}
\label{eq: loss_landscape}
    \begin{split}
        g_w(\alpha) & = \rho_w(\bm{W}+\alpha\bm{D}_w) = \mathcal{L}(\bm{W}+\alpha\bm{D}_w, \bm{X}), \\
        g_x(\alpha) & = \rho_x(\bm{X}+\alpha\bm{D}_x) = \mathcal{L}(\bm{W}, \bm{X}+\alpha\bm{D}_x),
    \end{split}
\end{equation}
where $\mathcal{L}$ is the training objective such as cross-entropy loss. $\bm{W}$ denotes the well-trained model parameters at the convergence of GNNs, and $\bm{D}_w$ denotes the random gradient directions sampled from Gaussian distribution. To facilitate the following expression and avoid confusion, we use $\bm{W}$ to represent model weights $\bm{W}^{(l)}$ from all the graph convolutional layers. The random gradient directions $\bm{D}_w$ are sampled and applied in the layer-wise fashion, and are scaled to have the same norm with $\bm{W}$ to eliminate the scaling invariance~\cite{visualloss} at each layer.
Similarly, $\bm{D}_x\in \mathbb{R}^{n\times d}$ is the random gradient direction possessing the same norm with node feature matrix $\bm{X}$. $\alpha$ is the scalar magnitude to denote the moving step size along the sampled directions.

\paragraph{\textbf{Sharp local minima in the loss landscapes}} To understand the model convergence behaviors of GNNs, we visualize the weight and loss landscapes of vanilla GCN trained on \emph{Pubmed} in Figure~\ref{fig:lossland} (a) and (b), respectively. First, in Figure~\ref{fig:lossland} (a), the weight loss landscape at the model convergence is much sharper than those of adversarial training methods, which we will introduce in the following to explicitly smooth the loss landscape. The flatness of loss surface is a popular metric to indicate the generalization ability: a flatter weight loss surface often shrinks the robustness generalization gap (i.e., the gap between training and testing performances)~\cite{gapsm, awp}. In other word, the converged model accompanied with larger loss curvature suffers from the poor trainability, and is more prone to overfitting. Being analog to other deep neural networks, the sharp weight minima of GNNs tends to damage the generalization performance in the out-of-distribution testing data. Second, for the feature loss landscape as shown in Figure~\ref{fig:lossland} (b), the cross-entropy loss changes significantly once the input features are slightly perturbed along the random sampled directions. The sharp feature minima indicates the overfitting on the seen node features, and makes the model vulnerable to the unseen adversarial perturbations. Considering the diverse node characteristics over the large-scale graph, such overfitting on the specific training subset is hard to generalize to the out-of-distribution testing data.

\paragraph{\textbf{The connection to generalization}} To further validate the strong correlation between the sharp local minima and the model's generalization performance in GNNs, we plot both the training dynamics and the generation gap in Figures~\ref{fig:lossland} (c) and (d), respectively. In Figure~\ref{fig:lossland} (c), the training dynamics are defined by the training and testing loss evolutions with the epochs. Although the training loss of vanilla GNNs decreases drastically, the testing loss is higher than those of adversarial training methods. This poor generalization performance is due to the sharp weight and feature minima, which overfit the training data (i.e., the lower training loss) and fail to extrapolate well to the testing data (i.e., the higher testing loss). In Figure~\ref{fig:lossland} (d), we directly visualize the generalization performance with the generalization gap, which is defined by the difference between the training and testing accuracy (i.e., the blue bar). It is observed that the generalization gap of vanilla GNNs is much larger than those of the adversarial training methods. Thus, the flatness of weight and feature loss landscapes are strongly correlated with the
generalization performance in GNNs. At the model convergence, the sharp loss surfaces in terms of weights and features will weaken the generalization performance in the unseen testing data.

\section{Adversarial Training of GNNs}
In this section, we start with introducing the traditional adversarial training methods, which  leverage adversarial perturbations in the model weights or input features. To smooth both the weight and feature loss landscapes in training GNNs, we propose a new training approach, named co-adversarial perturbation (CAP), to relieve the overfitting issue and improve the generalization ability to the maximum extent.

\subsection{Single-type Adversarial Training}
\paragraph{\textbf{Adversarial weight perturbation}} Motivated by the close connection between the flatness of weight loss landscape and the model's generalization performance, AWP \cite{awp} has been proposed to smooth the loss surface curvature in the domain of computer vision modeling. To be specific, AWP aims to minimize the worst-case loss within the small region centered at the model weights, which could be formulated as a minimax objective:
\begin{equation}
    \label{eq: awp}
    \min_{\bm{W}} \max_{||\bm{\epsilon}_w||_p\leq \rho} \mathcal{L}(\bm{W}+\bm{\epsilon}_w, \bm{X}),
\end{equation}
where $\bm{\epsilon}_w$ is the crafted adversarial perturbation, which is constrained
within a $l_p$ norm ball centered at $\bm{W}$ with radius $\rho$. $||\cdot||_p$ is the p-norm distance metric. Intuitively, AWP seeks to obtain parameter $\bm{W}$ whose neighborhood regions have low training loss. In this way, the weight loss landscape has smaller curvature at the final learned weights, which in turn shrinks the generalization gap.

\paragraph{\textbf{Adversarial feature perturbation}} Deep neural networks often fall vulnerable when it is presented with adversarial examples crafted with imperceptible perturbations, and make the wrong predictions on them. Among the series of defensive approaches, the adversarial training based on  feature perturbation generally has achieved superior robustness than others. In contrary, a few recent studies turn to ameliorate networks’ generalization ability via the adversarial training~\cite{vlr, afan, freelb}. Specifically, instead of training with original samples, it trains neural networks by incorporating the adversarial feature perturbations on the inputs as follows:
\begin{equation}
    \label{eq: afp}
    \min_{\bm{W}} \max_{||\bm{\epsilon}_x||_p\leq \rho} \mathcal{L}(\bm{W}, \bm{X}+\bm{\epsilon}_x),
\end{equation}
where $\bm{\epsilon}_x$ denotes the adversarial feature perturbation bounded by the $l_p$ norm ball. Similar to the above training method based on adversarial weight perturbation, the optimization of Eq.~\eqref{eq: afp} seeks to find the optimal weights having the smaller losses at the neighborhood regions centered at the input features. Therefore, the feature loss landscape is expected to be smoothed to reduce the generalization gap. The maliciously perturbed features could also be treated as the augmentation data, which helps model extrapolate the out-of-distribution testing data and alleviate the overfitting.

\subsection{Co-adversarial Perturbation Training}\label{subsec:cap}
Although the adversarial training has been extensively investigated in the computer vision and language models, the in-depth exploration of adversarial training to improve the generalization ability of GNNs is still under-explored. Considering the graph analytics, almost all of the adversarial-training-based methods concentrate in defending the adversarial attacks~\cite{attreview}. Although the model robustness of GNNs is substantially enhanced, the adversarial training usually compromises the standard accuracy on the clean graph data~\cite{hurt, odds}. This leads to the open question: how the adversarial training could be used effectively to ameliorate the generalization performance of GNNs in the graph analytics?

However, it is non-trivial to directly extend the existing adversarial training methods to GNNs due to the following two challenges. First, the generalization ability of GNNs is constrained by the sharp local minima in both the weight and feature loss landscapes. In contrast, the off-the-shelf adversarial training leverages either the adversarial weight or feature perturbation to smooth one of the loss landscapes. This could limit the maximum potential of adversarial training in boosting the standard classification accuracy, and may lead to the opposite answer for the above research question. Second, to simultaneously alleviate the sharpness of the two landscapes in GNNs, the exact computation of couple weight and feature perturbations is extremely time-consuming. Considering the minmax framework of adversarial training, the co-optimization of feature and weight perturbations in the inner maximization loop makes the solution space too large to be explored efficiently.

To tackle these two challenges, we propose a simple yet effective graph adversarial training algorithm, named CAP. We first mathematically formulate the co-adversarial training problem, and then state how CAP iteratively injects the weight and feature perturbations and  controls the adversarial frequency.

\begin{algorithm}
    \caption{CAP: Co-Adversarial Perturbation}\label{cap}
    \KwIn{Graph $\mathcal{G}=(\mathcal{V}, \mathcal{E})$; input feature matrix $\bm{X}$; learning rate $\tau$; training epochs $N$; total time step $T$; $\mathcal{L}$ as the loss function for the GNN model with model parameters $\bm{W}$ on graph $\mathcal{G}$.}

    \KwOut{Optimized model parameters $\bm{W}$.}

    \tcc{Stage 1: the standard training}
    \For(){$e = 1 \dots S$}{
        Performing the vanilla training process \;
    }

    \tcc{Stage 2: the CAP training}
    \For(){$e = (S+1) \dots N$}{
        \eIf(\tcc*[f]{Adversarial Feature Perturbation}){$e \% F == 0$}{
            Calculate $\bm{\epsilon}_x^T$ with Eq.~\eqref{eq: afp} \;
            $ L_{adv} = \mathcal{L}(\bm{W}, \bm{X}+\bm{\epsilon}_x^T) $ \tcc*{Calculate the perturbed loss}

        }(\tcc*[f]{Adversarial Weight Perturbation}){
            Calculate $\bm{\epsilon}_w^T$ with Eq.~\eqref{eq: awp} \;
            $L_{adv} = \mathcal{L}(\bm{W} + \bm{\epsilon}_w^T, \bm{X})$ \tcc*{Calculate the perturbed loss}
        }
        $\bm{W} = \bm{W} - \tau \cdot \nabla_{\bm{W}}L_{adv} $ \tcc*{Optimize model parameters}
    }
    \KwRet{$\bm{W}$} \;
\end{algorithm}

\paragraph{\textbf{Co-adversarial training problem}} Instead of incorporating the single  perturbation during model training in Eq.~\eqref{eq: awp} or \eqref{eq: afp}, the co-adversarial training targets at seeking the couple weight and feature perturbations, along which the training loss increases dramatically. Mathematically, the minmax objective of co-adversarial training is:
\begin{equation}
    \label{eq: cap}
    \min_{\bm{W}} \max_{||\bm{\epsilon}_w||_p\leq \rho, ||\bm{\epsilon}_x||_p\leq \rho} \mathcal{L}(\bm{W}+\bm{\epsilon}_w, \bm{X}+\bm{\epsilon}_x).
\end{equation}
According to the above optimization problem, the inner optimization generates the  adversarial perturbations in the union spaces of weights and features, and  smooths both the weight and feature loss landscapes simultaneously. Unfortunately, due to the hugeness of union spaces, it is intractable to obtain the optimal pair of weight and feature perturbations. To efficiently conduct the co-adversarial training in GNNs, we propose CAP functioning with three key components: alternative adversarial  perturbations, alternative frequency adjustment, and two-stage training scheme.

\paragraph{\textbf{Alternative adversarial perturbation in CAP}} The spirit of CAP is to break the couple perturbations in the inner loop of Eq.~\eqref{eq: cap} into two independent phases. Each of them applies the adversarial weight or feature perturbation, and alternatively updates GNNs.

To be specific, in the training phase of adversarial weight perturbation, the co-adversarial training in Eq.~\eqref{eq: cap} is simplified to the optimization problem in Eq.~\eqref{eq: awp}.  Since
the exact solution of the inner maximization $\bm{\epsilon}^*_w = \mathrm{argmax}_{||\bm{\epsilon}_w||_p\leq \rho} \mathcal{L}(\bm{W}+\bm{\epsilon}_w, \bm{X})$ is intractable, we adopt the multi-step projected gradient descent (PGD) to approximate it. At step $t+1$, the approximation of weight perturbation is given by:
\begin{equation}
\label{eq:pgd_w}
   \bm{\epsilon}^{t+1}_w = \Pi_{\rho}\left(\bm{\epsilon}^{t}_w + \beta \cdot \frac{\nabla_{w}\mathcal{L}(\bm{W}+\bm{\epsilon}^t_w, \bm{X})}{\Vert \nabla_{w}\mathcal{L}(\bm{W}+\bm{\epsilon}^t_w, \bm{X}) \Vert_p} \right).
\end{equation}
$\Pi_{\rho}$ is a
projection function projects the computed perturbation back to the surface of $l_p$ norm ball if the perturbation is out of the ball, and $\beta$ is the step size of the inner maximization. Note that $\bm{\epsilon}^0_w$ is zero at the beginning of PGD. Let $T$ denotes the total time step, where we could obtain the  approximated weight perturbation  $\bm{\epsilon}^T_w$. The training of GNN models is then conducted by minimizing the standard cross-entropy objective: $ \min_{\bm{W}}  \mathcal{L}(\bm{W}+\bm{\epsilon}^T_w, \bm{X})$.

In the training phase of adversarial feature perturbation, the co-adversarial training in Eq.~\eqref{eq: cap} is simplified to the optimization problem in Eq.~\eqref{eq: afp}. Similarly, we adopt PGD to approximate the worst-case feature perturbation in the inner maximization loop. The iterative adversarial feature perturbation at time step $t$ is:
\begin{equation}
    \label{eq: pgd_f}
    \bm{\epsilon}^{t+1}_x = \Pi_{\rho}(\bm{\epsilon}^{t}_x + \beta\cdot\mathrm{sign}(\nabla_{x}\mathcal{L}(\bm{W}, \bm{X}+\bm{\epsilon}^t_x))),
\end{equation}
where $\mathrm{sign}(\cdot)$ is the sign function. After $T$ steps, the approximated feature perturbation is applied to augment the node features and train GNNs: $\min_{\bm{W}}  \mathcal{L}(\bm{W}, \bm{X}+\bm{\epsilon}^T_x)$.

\paragraph{\textbf{Alternative frequency adjustment in CAP}}
As aforementioned, the alternative training conducts one of the perturbation methods at each epoch to update GNN models. On the one hand, the adversarial weight perturbation can influence the losses of all nodes, and considers the global model-wise generalization improvement. On the other hand, the adversarial feature perturbation augments each node with the worst-case loss direction, which could be regarded as the local example-wise perturbation that does not consider the overall
effect on other nodes. Given the diverse node characteristics over the graph, the model trained to adapt to specific node perturbation may cannot well extrapolate the  other nodes. Furthermore, the set of adversarially generated nodes is potential to mismatch the ground-truth distribution in the large-scale graph. Such diverse feature perturbations and the mismatched distribution will hurt the classification accuracy on the original clean data. This is also why the traditional adversarial training in GNNs commonly comprises the generalization performance to obtain the better robustness.

To control the potential negative impact brought from the adversarial feature perturbation, we alternatively conduct the two perturbation methods with frequency of $F$. For every continuous time frame with $F$ training epochs, we adopt the adversarial weight perturbation at the first $F-1$ epochs, following which the adversarial feature perturbation is applied. We use a lager $F$ to properly involve the adversarial feature perturbation to smooth the feature loss landscape, while avoiding the accuracy dropping due to the overly-mismatched generated node features.

\begin{table*}[t]
\caption{Dataset statistics (“m” stands for \textbf{m}ulti-class classification, and “s” for \textbf{s}ingle-class.)}
\label{stats}
\centering
\begin{tabular}{cccccc}
\toprule
Dataset & Nodes & Edges & Features & Classes & Train/Val/Test \\ \midrule\midrule
ogbn-arxiv & 169,343 & 1,166,243 & 128 & 40(s) & 0.55 / 0.15 / 0.30 \\
ogbn-proteins & 132,534 & 39,561,252 & 0 & 112(m) & 0.65 / 0.15 / 0.20 \\
Pubmed & 19,717 & 44,338 & 500 & 3(s) & 0.60 / 0.20 / 0.20 \\
Yelp & 716,847 & 6,977,410 & 300 & 100(m) & 0.75 / 0.10 / 0.15 \\
ogbn-products & 2,449,029 & 61,859,140 & 100 & 47(s) & 0.08 / 0.02 / 0.90 \\
\bottomrule
\end{tabular}
\end{table*}

\paragraph{\textbf{Two-stage training in CAP}}
One common phenomenon in training GNNs is that the cross-entropy loss is decreased rapidly at the initial training stage. In other words, the model weights are iterating to the low-loss regions, and far away from falling into the sharp local minima in the loss landscape. However, the adversarial weight and feature perturbations will impair the convergence speed at the initial stage, and may even mislead GNNs towards the worse model spaces. This problem encourages us to propose the generic two-stage training strategy: the first stage of $S$ epochs trains GNNs standardly, and the second stage adversarially trains GNNs with the remaining epochs. It is more rational to regularize GNN training at the second stage, wherein the model at convergence tends to fall into the sharp local minima.

\subsection{Generalization Analysis of Adversarial Training}
In this section, we analyze how our CAP smooth the loss landscapes to shrink the generalization gap. To our best knowledge, the only effort conjoining GNNs and adversarial training to improve model's generalization ability is FLAG~\cite{flag}, which is unofficially published in ArXiv. FLAG directly adopts the single feature perturbation to augment clean data, and misses the detailed analysis on how the adversarial training works in GNNs.

In Figure~\ref{fig:lossland}, we apply CAP and FLAG on vanilla GCN, and compare them in terms of the flatness of loss landscapes and the generalization gap. Comparing with vanilla training and FLAG,
our CAP can obtain the smoothest weight and feature loss landscapes. Notably, this brings CAP the smallest generalization gap (i.e., blue bar) as shown in  Figure~\ref{fig:lossland} (d), and delivers the highest test accuracy. The overfitting issue is also greatly relieved by CAP, which is empirically demonstrated by the smallest gap between the training and testing losses in Figure~\ref{fig:lossland} (c). These results positively provide the answer to our research question--the adversarial training based on weight and feature perturbations could improve the  generalization ability of GNNs to the maximum extent. Although FLAG also shows better results on \emph{Pubmed}, we will explain in the following  experiments that the single feature perturbation is limited in generalizing models and even damaging them in some applications.

\section{Experiments}
In this section, we conduct extensive experiments on various datasets to empirically demonstrate CAP’s effectiveness through answering these following questions.
\begin{itemize}
    \item \textbf{Q1:} How effective is the proposed CAP applied to current popular GNNs in the  node classification tasks?
    \item \textbf{Q2:} Is CAP generally applicable to the different mini-batch training algorithms on the large-scale graphs?
    \item \textbf{Q3:} How does each component of CAP affect the model performance?
    \item \textbf{Q4:} How robust CAP is in the poisoned attack data?
\end{itemize}
\subsection{Benchmark Datasets}
Our model is evaluated  on five real-world benchmark datasets:  \emph{ogbn-arxiv}, \emph{Pubmed}, \emph{ogbn-proteins}, \emph{Yelp} and \emph{ogbn-products}. They range from the small to large scales, to comprehensively validate the generalization behaviors of adversarial training. The basic information and each dataset division are shown in Table~\ref{stats}. More details are reported in Appendix \ref{app:A}.

\subsection{Baselines}
We compare the vanilla training, FLAG and our CAP over different GNN backbones in terms of their generalization test accuracies.
Specifically, we run them on  three prestigious backbones, namely GCN~\cite{gcn}, GraphSAGE~\cite{graphsage}, and GAT~\cite{gat}. The detailed description of these these backbones are in Appendix~\ref{sec: backbone_des}.

\subsection{Experimental Settings} For \emph{ogbn-arxiv}, \emph{Pubmed}, and \emph{ogbn-proteins},u we use test accuracy as the evaluation metric while test ROC-AUC score for \emph{ogbn-proteins}, \emph{Yelp}. Following common practice, we report the test performance associated with the best validation result. For a fair comparison, we conduct ten independent runs and report the mean result with the standard deviation over it. For \emph{ogbn-arxiv}, \emph{ogbn-proteins}, \emph{Yelp}, and \emph{ogbn-products}, we use the standard split for our experiments. Since \emph{Pubmed}  does not provide a standard split for the full-supervised learning, we randomly split the nodes of each class into 60\%, 20\%, and 20\% for training, validation and testing, and measure the performance of all models on the test sets over ten random splits, as suggested in \cite{pubmedsplit}.

\begin{table*}[]
\caption{Test performance in percent on \emph{ogbn-arxiv}, \emph{ogbn-proteins}, \emph{Pubmed} and \emph{Yelp} datasets. The blanks are due to the bad scalability of backbone models in the concerned datasets. The highest results are in bold.}
\label{tab:performance}
\centering
\begin{tabular}{cccccccc}
\toprule
Backbones & Varients & \begin{tabular}[c]{@{}c@{}}ogbn-arxiv\\ Test Acc\end{tabular} & \begin{tabular}[c]{@{}c@{}}ogbn-proteins\\ Test ROC-AUC\end{tabular} & \begin{tabular}[c]{@{}c@{}}Pubmed\\ Test Acc\end{tabular} & \begin{tabular}[c]{@{}c@{}}Yelp \\ Test ROC-AUC\end{tabular} & \begin{tabular}[c]{@{}c@{}}ogbn-products\\ Test Acc\end{tabular} & \begin{tabular}[c]{@{}c@{}}Avg. \\ improvement\end{tabular} \\ \midrule \midrule
\multicolumn{1}{c}{\multirow{3}{*}{GCN}} & \ Vanilla\  & 71.74±0.29 & 72.51±0.35 & 88.14±0.27 & - & - & 1.22 \\
\multicolumn{1}{c}{} & \ +FLAG\  & 71.90±0.25 & 71.71±0.50 & 88.92±0.47 & - & - & 1.23 \\
\multicolumn{1}{c}{} & \ +CAP\  & \textbf{72.12±0.24} & \textbf{73.50±1.16} & \textbf{89.70±0.43} & - & - & - \\ \midrule
\multicolumn{1}{c}{\multirow{3}{*}{GraphSAGE}} & \ Vanilla\  & 71.49±0.27 & 76.91±1.02 & 89.32±0.42 & 87.89±0.15 & 78.70±0.36 & 1.02 \\
\multicolumn{1}{c}{} & \ +FLAG\  & 72.19±0.21 & 76.44±0.81 & 89.91±0.36 & 88.82±0.16 & 79.09±0.42 & 0.50 \\
\multicolumn{1}{c}{} & \ +CAP\  & \textbf{72.34±0.22} & \textbf{77.53±0.70} & \textbf{90.26±0.26} & \textbf{88.97±0.17} & \textbf{79.33±0.31} & - \\ \midrule
\multicolumn{1}{c}{\multirow{3}{*}{GAT}} & \ Vanilla\  & 73.65±0.11 & - & 90.10±0.29 & 76.80±0.19 & 79.45±0.59 & 1.06 \\
\multicolumn{1}{c}{} & \ +FLAG\  & 73.66±0.17 & - & 89.56±0.31 & 77.38±0.42 & 81.55±0.35 & 0.35 \\
\multicolumn{1}{c}{} & \ +CAP\  & \textbf{73.72±0.16} & - & \textbf{90.17±0.32} & \textbf{77.56±0.49} & \textbf{81.89±0.35} & - \\
\bottomrule
\end{tabular}
\end{table*}

Our backbone network implementations are derived from publicly released code without making any changes to the model architecture or training setup. We do full-batch training on \emph{ogbn-arxiv}, \emph{ogbn-proteins} and \emph{Pubmed}, while adopt neighbor sampling \cite{graphsage} as the mini-batch algorithm on \emph{Yelp} and \emph{ogbn-products} to make the experiments scalable.
For simplicity, we always use three time steps($T=3$), either in the feature perturbation or the weight perturbation. We tune the following hyper-parameters:
(1) the number of training epochs in the first stage $T \in \{ 5, 10, 20, 50, 100, 200\}$,
(2) frequency of the adversarial feature perturbation $F \in \{2, 5, 10, 20\}$,
(3) step size $\beta \in [1e-4, 1e-2]$.

\subsection{Overall Results}
To answer the research question \textbf{Q1}, we summarize the results of full-supervised node classification tasks in Table~\ref{tab:performance}. After analyzing the results, we have some observations.

First, we can observe that our proposed CAP achieves improvement of test performance over all datasets, which demonstrate the effectiveness of our method. 
The average improvement obtained by CAP could reach up to $0.9\%$.
Second, the feature perturbation can improve generalization in the large-scale graph. On the \emph{Yelp} dataset, the backbones equip with the FLAG technique all receive promising results, improving the performance of GraphSAGE and GAT by 1.06\% and 0.76\% respectively. The \emph{Yelp} dataset is a large-scale graph possessing over 700,000 nodes, with massive out-of-distribution test nodes. The distribution gap causes the vanilla models are prone to overfitting to the training subset and lack generalization ability to unseen testing data. 
With the addition of adversarial feature perturbations, the adversarial samples can be deemed as augmentation data to improve the generalization ability of models. Notably, when CAP is applied, the backbone network obtain a further performance improvement. This shows that the model earns global model-wise generalization improvement by the weight perturbation. Similar results are found on the \emph{ogbn-arxiv} and \emph{ogbn-products} datasets. The above-mentioned observations illustrate that CAP can maximize the generalization ability of the model by smoothing the feature and weight loss landscapes jointly compared to FLAG.

Besides, we observe that FLAG reports three worse results in all twelve experimental setups. Especially on the \emph{ogbn-proteins} dataset, FLAG compromises an average of 0.64\% on the test ROC-AUC score of all backbone networks. The \emph{ogbn-proteins} is a small graph in Open Graph Benchmark and is markedly denser than the other graphs. This leads to the fact that even if only slight perturbations are injected into input features, they can significantly affect the representation of nodes after aggregation and propagation. Meanwhile, FLAG performs feature perturbations in each epoch, which encourages the model to fit adversarial samples and weakens the classification ability of clean data. The design of alternative frequency adjustment in our method solves this problem well. By reducing the frequency of the feature perturbation, CAP can avoid its potential negative impact. The improvement of 2.50\% over GCN equips with FLAG demonstrates the superiority of our CAP.

In summary, our CAP achieves superior performance on all datasets. The results verify the superiority of perturbing the model weights and node features jointly, which makes the model avoid falling into sharp local minima and obtains better generalization.

\begin{table}[]
\centering
\caption{Test ROC-AUC score in percent on \emph{Yelp} with different mini-batch sub-graph sampling methods.}
\label{tab:sampling}
\begin{tabular}{clcc}
\toprule
\begin{tabular}[c]{@{}c@{}}Sampling\\ Methods\end{tabular} & Variants & \begin{tabular}[c]{@{}c@{}}Yelp\\ Test ROC-AUC\end{tabular} & \begin{tabular}[c]{@{}c@{}}Avg.\\ improvement\end{tabular} \\ \midrule \midrule
\multirow{3}{*}{\begin{tabular}[c]{@{}c@{}}GraphSAGE\\ w/ NS\end{tabular}} & Vanilla & 87.89±0.15 & 1.23 \\
 & +FLAG & 88.82±0.16 & 0.17 \\
 & +CAP & 88.97±0.17 & - \\ \midrule
\multirow{3}{*}{\begin{tabular}[c]{@{}c@{}}GraphSAGE\\ w/ Cluster\end{tabular}} & Vanilla & 76.73±0.09 & 0.61 \\
 & +FLAG & 76.92±0.17 & 0.36 \\
 & +CAP & 77.20±0.23 & - \\ \midrule
\multirow{3}{*}{\begin{tabular}[c]{@{}c@{}}GraphSAGE\\ w/ SAINT\end{tabular}} & Vanilla & 75.26±0.14 & 0.61 \\
 & +FLAG & 75.63±0.20 & 0.12 \\
 & +CAP & 75.72±0.09 & - \\ \bottomrule
\end{tabular}
\end{table}

\subsection{Generality with Mini-batch Methods}
The success of GNNs in large-scale graph is inseparable from the mini-batch sub-graph sampling methods. We evaluate our CAP method with the different sampling methods on \emph{Yelp}, to illustrate the generality of our method and answer research question \textbf{Q2}. We adopt Neighbor Sampling~\cite{graphsage}, Cluster sampling~\cite{clustergcn} and SAINT sampling~\cite{yelp} as three samplers working on GraphSAGE backbone network. The results are shown in Table \ref{tab:sampling}. After CAP intervenes, the performance on the test set rises by an average of 0.82\%. Such results further show the effectiveness of our method on large-scale graphs and the compatibility with different sampling methods.

\subsection{Ablation Study}
 In this part, we delve into CAP to investigate its each component and answer the research question \textbf{Q3}. To illustrate the effectiveness of the co-adversarial perturbation, we respectively replace the feature perturbation and the weight perturbation with vanilla training procedures, and evaluate the performance on various datasets. We choose GCN and GraphSAGE as backbone networks and conduct ablation experiments on \emph{ogbn-arxiv} and \emph{ogbn-proteins}. The variant "+WP" indicates that the original training procedure is used to replace the epoch of the feature perturbation introduced in Section \ref{subsec:cap}. The variant "+FP" is also similar. Table \ref{tab:ablation} shows the test performance for different ablation variants. We can see that the absence of any kind of perturbation causes the model to perform worse on the test set. This proves the effectiveness of our co-adversarial perturbation method. We also find that different datasets are differently sensitive to feature perturbations and weight perturbations. For example, applying feature perturbations on the \emph{ogbn-arxiv} dataset has more significant effects while weight perturbations are more powerful on the \emph{ogbn-proteins} dataset.

\begin{table}[]
\setlength{\tabcolsep}{2.5pt}
\caption{Ablation study of the feature perturbation and the weight perturbation.}
\label{tab:ablation}
\centering
\begin{tabular}{clccc}
\toprule
 Backbones & Variants & \begin{tabular}[c]{@{}c@{}}ogbn-arxiv\\ Test Acc\end{tabular} & \begin{tabular}[c]{@{}c@{}}ogbn-proteins\\ Test ROC-AUC\end{tabular} & \begin{tabular}[c]{@{}c@{}}Pubmed\\ Test Acc\end{tabular} \\ \midrule \midrule
\multirow{3}{*}{GCN} & +CAP & 72.12±0.24 & 73.50±1.16 & 89.70±0.43 \\
 & +WP & 71.79±0.21 & 72.72±1.25 & 89.00±0.38 \\
 & +FP & 71.94±0.30 & 72.05±1.04 & 89.35±0.44 \\ \midrule
\multirow{3}{*}{GraphSAGE} & +CAP & 72.34±0.22 & 77.53±0.70 & 90.26±0.26 \\
 & +WP & 72.31±0.21 & 76.74±0.41 & 89.28±0.35 \\
 & +FP & 72.30±0.20 & 76.53±0.93 & 89.91±0.35 \\ \bottomrule
\end{tabular}
\end{table}

\begin{table}[]
\caption{Test accuracy on \emph{Pubmed} with the different skip epochs in the two-stage training of CAP.}
\label{tab:skipacc}
\begin{tabular}{ccccc}
\toprule
\multicolumn{1}{c}{\multirow{2}{*}{Baselines}} & \multicolumn{4}{c}{Skip epochs $S$} \\ \cmidrule(l){2-5}
\multicolumn{1}{c}{}                           & 0   & 100  & 200  & 400  \\ \midrule \midrule
GCN + CAP                                      & 89.45\% & 90.55\% & 90.06\% & 89.40\% \\ \bottomrule
\end{tabular}
\end{table}

\begin{figure}[t]
    \centering
    \includegraphics[width=0.45\textwidth]{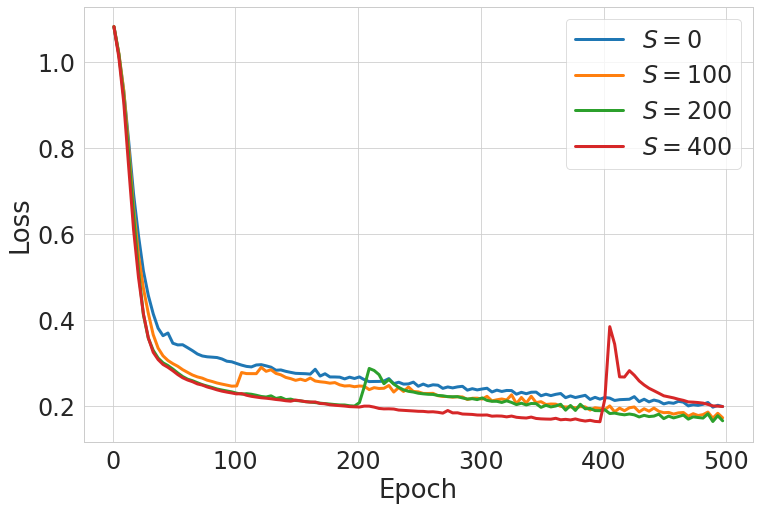}
    \caption{Loss curves of different skip epochs $S$ on GCN backbone and \emph{Pubmed}.}
    \label{fig:skip}
\end{figure}

\subsection{Analysis of Two-stage Training in CAP}
We introduce the two-stage training scheme in Section \ref{subsec:cap} and briefly analyze its impact on the model convergence. Furthermore, we are curious about how the two-stage training scheme affects the training process, and further answer research question \textbf{Q3}. We compare the impacts of different skip epochs $S$, which is applied in the first stage to conduct the pure vanilla training. When the total number of epochs is 500, we choose four different values [0, 100, 200, 400] for skip epochs $S$ while ensure that other hyperparameters are consistent, and observe the trend of training loss. As the Figure \ref{fig:skip} shows, reasonable setting of $S$ can help the model converge better. When $S=0$, on account of the model is perturbed in the initial stage of training, the model converges where the training loss is higher than other settings.
Meanwhile, it can easily observed that the more appropriate choices for $S$ are 100 and 200. The best accuracy can reaching 90.55\% on the test set when $S$ is 100, but at 400 the final training loss is gradually tend to be similar with the situation "$S=0$".
Furthermore, we notice that as the value of $S$ increases, the training loss grows simultaneously in the first epoch during the adversarial training.
This shows that the vanilla training causes the model more easily converge to the sharp local minima which makes the model more susceptible to the impact of adversarial perturbations. Table \ref{tab:skipacc} shows the testing accuracy rates for different values of $S$. When $S$ is 0 or 400, the testing accuracy moves down significantly with which is compared the other two cases. It is in line with our analysis that a suitable $S$ can obviously help the model converge better, and improve the performance of the model as well.

\begin{figure}[t]
    \includegraphics[width=.45\textwidth]{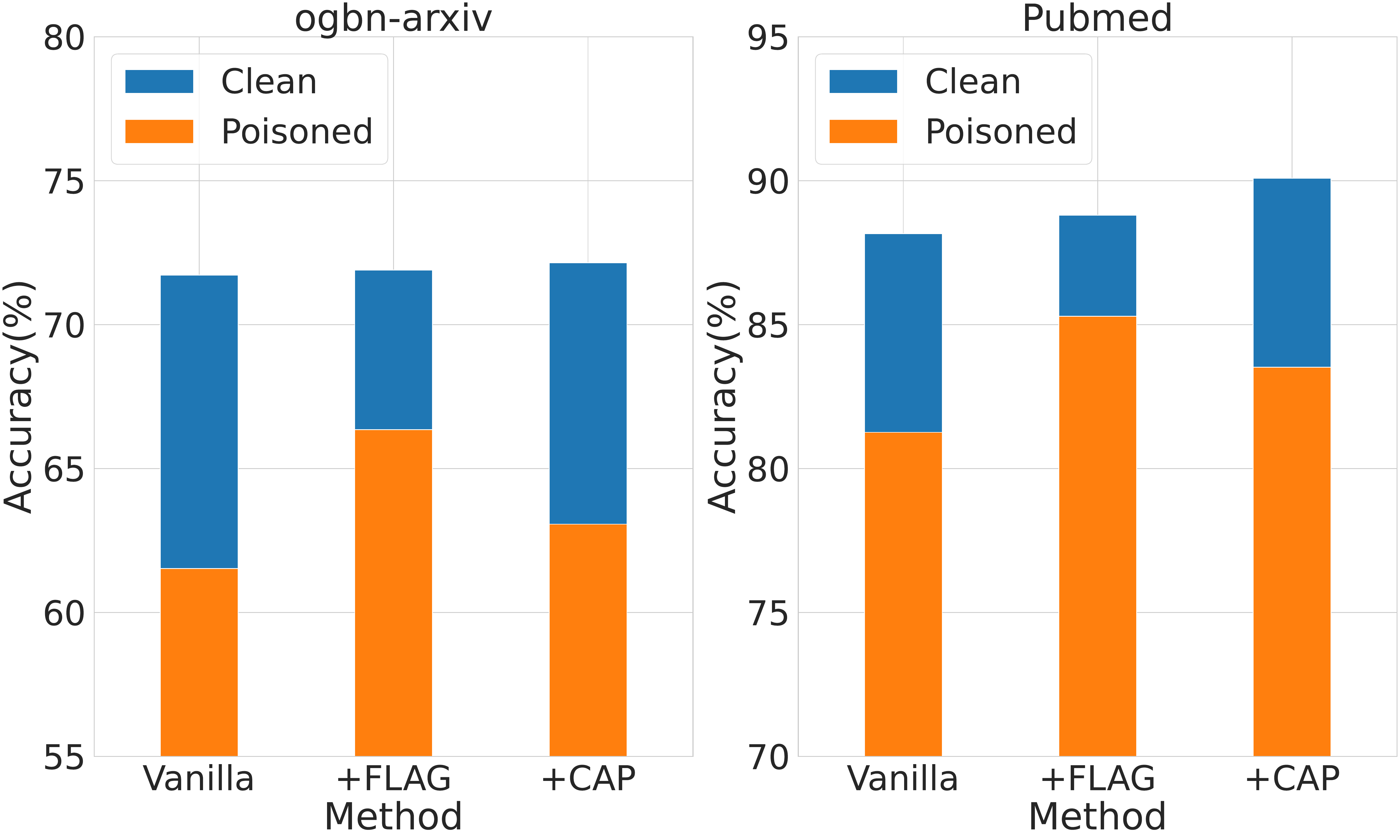}
    \caption{Test accuracy with and without adversarial attack on \emph{ogbn-arxiv} and \emph{Pubmed}.}
    \label{fig:att}
\end{figure}

\subsection{Adversarial Attacks}
Despite our proposed CAP primarily focuses on improving the generalization of GNNs, we wonder whether the model simultaneously has a measure of defense against attacks owing to perturbations crafted by the adversarial attack method. We conduct evasion attacks on input features, and compare the robustness to answer \textbf{Q4}. Specifically, we perform the training process on clean data, and finally test using the input features with Gaussian noise added. We use GCN as the victim model and conduct experiments on \emph{ogbn-arxiv} and \emph{Pubmed}. As is shown in Figure~\ref{fig:att}, FLAG achieves a higher testing accuracy on the poisoned data than the other methods after being attacked. In other words, FLAG has better robustness against the adversarial attack of input features. By incorporating adversarial feature perturbations, FLAG minimizes the node-wise worst loss, thereby defending against random noise attacks. Surprisingly, although our CAP method is not as good as FLAG, it still improves the robustness of vanilla models partly. Intergrating the testing performance shown in Table \ref{tab:performance}, we believe that there is a trade-off between generalization ability and robustness.

\section{Conclusion}
In this paper, we first study the weight and feature loss landscapes for GNNs, and clarify the close connection between the sharp local minima at the loss landscapes and the poor generalization performances. Based upon the loss analysis, we shred new insight to the design of adversarial training in GNNs. Specifically, we formulate the co-adversarial training problem targeting at flatten the feature and weight loss landscapes. To efficiently optimize the problem, we propose CAP to alternatively conduct the weight and feature perturbations with appropriate alternative frequency. The two-stage training strategy is further leveraged to ensure the quick convergence of GNNs at the initial training stage. Extensive experiments show that our proposed CAP could generally smooth the loss landscapes and shrink the generalization gap, which in turn improves the test  performance in the large-scale graph analytics.

\bibliographystyle{ACM-Reference-Format}
\bibliography{sample-base}

\clearpage
\appendix
\section{Appendix}
\subsection{Datasets} \label{app:A}
Our model is evaluated on five real-world datasets including two citation networks, two social networks and a protein-protein interaction network.
\begin{itemize}
    \item \emph{ogbn-arxiv}~\cite{ogb}. The \emph{ogbn-arxiv} dataset is a directed graph with nodes representing Computer Science (CS) arXiv papers and edges representing the citation relationship between papers. The node feature is obtained by averaging the embedding of words in the title and abstract.
    \item \emph{Pubmed}~\cite{pubmed}. The \emph{Pubmed} dataset consists of 19,717 publications from PubMed database as nodes. The edges indicate reference relationships. The difference is that each publication in dataset is described by a TF/IDF weighted word vector from a dictionary which consists of 500 unique words.
    \item \emph{ogbn-proteins}\cite{ogb}.
    The \emph{ogbn-proteins} dataset is an undirected graph. Nodes represent proteins, while edges denote various biologically significant relationships between proteins, such as physical interactions, co-expression or homology~\cite{prot1, prot2}.
    \item \emph{Yelp}~\cite{yelp}. The \emph{Yelp} dataset is a social network prepared from the open challenge website\footnote{https://www.yelp.com/dataset}. Each node denotes a user, and each edge represents a friendship between two users. The node features are from averaging the embedding of words in all his review. The categories of the businesses reviewed by a user as the multi-class label of him.
    \item \emph{ogbn-products}~\cite{ogb}. The \emph{ogbn-products} dataset is an Amazon product co-purchasing network. Nodes are products sold in Amazon, and edges between two products indicate that the products are purchased together. The process of node features and labels follows \citet{clustergcn}.
\end{itemize}

\subsection{Backbone Description}
\label{sec: backbone_des}
We run the different training approaches on three prestigious backbones, namely GCN~\cite{gcn}, GraphSAGE~\cite{graphsage}, and GAT~\cite{gat}. They are described as below.
\begin{itemize}
    \item \textbf{GCN}\cite{gcn}.
    GCN is the first notable GNN research on graph embedding. To learn graph embedding, GCN adopts neighborhood aggregation and feature transformation functions to encode both local graph connectivity structure and features of nodes.
    GCN can learn representation for a central node by  aggregating representations of the neighbors iteratively.
    \item \textbf{GraphSAGE}\cite{graphsage}. GraphSAGE extends the GCN into an inductive framework by training a set of aggregator functions to generate embeddings for each node. It is a general inductive framework which generates embeddings by sampling and aggregating features from a node’s local neighborhood.
    Most previous approaches are inherently transductive, and they require that all nodes in the graph are present during the training of the embeddings. Therefore, these methods do not have the ability to generate embeddings for unseen nodes. In contrary, GraphSAGE is capable of encoding the unseen nodes with the seen neighborhood embeddings.
    \item \textbf{GAT}\cite{gat}.  GCN gathers information from local neighborhood and all neighbors contribute equally in the message passing. To overcome the issue, GAT is proposed. GAT incorporates the attention mechanism into the propagation step and learns the edge weights at each layer based on node features. Instead of considering all neighbors with equal importance, as is done in GCN, GAT learns to give varied levels of weights to nodes in each node's neighborhood, and the weights are adaptively learned during the model training.
\end{itemize}








\end{document}